\documentclass{acmproc-sp}

\usepackage[table]{xcolor}
\usepackage{enumitem}

\usepackage[
  pass,
]{geometry}

\begin{document}

\title{OpenML: networked science in machine learning}

\numberofauthors{1} 
\author{\alignauthor Joaquin Vanschoren\textsuperscript{1}, Jan N. van Rijn\textsuperscript{2}, Bernd Bischl\textsuperscript{3}, and Luis Torgo\textsuperscript{4} \\
\affaddr{\textsuperscript{1} Eindhoven University of Technology -  j.vanschoren@tue.nl}\\
\affaddr{\textsuperscript{2} Leiden University -  j.n.van.rijn@liacs.leidenuniv.nl}\\
\affaddr{\textsuperscript{3} Technische Universit\"{a}t Dortmund -  bischl@statistik.tu-dortmund.de}\\
\affaddr{\textsuperscript{4} INESC Tec / University of Porto -  ltorgo@dcc.fc.up.pt}
}

\date{8 May 2014}
\maketitle
\begin{abstract}
Many sciences have made significant breakthroughs by adopting online tools that help organize, structure and mine information that is too detailed to be printed in journals. In this paper, we introduce OpenML, a place for machine learning researchers to share and organize data in fine detail, so that they can work more effectively, be more visible, and collaborate with others to tackle harder problems. We discuss how OpenML relates to other examples of networked science and what benefits it brings for machine learning research, individual scientists, as well as students and practitioners.
\end{abstract}

\section{Introduction}
When Galileo Galilei discovered the rings of Saturn, he did not write a scientific paper. Instead, he wrote his discovery down, jumbled the letters into an anagram, and sent it to his fellow astronomers. This was common practice among respected scientists of the age, including Leonardo, Huygens and Hooke.

The reason was not technological. The printing press was well in use those days and the first scientific journals already existed. Rather, there was little personal gain in letting your rivals know what you were doing. The anagrams ensured that the original discoverer alone could build on his ideas, at least until someone else made the same discovery and the solution to the anagram had to be published in order to claim priority.

This behavior changed gradually in the late 17th century. Members of the Royal Society realized that this secrecy was holding them back, and that if they all agreed to publish their findings openly, they would all do better \cite{kealey2010sex}. Under the motto ``take nobody's word for it'', they established that scientists could only claim a discovery if they published it first, if they detailed their experimental methods so that results could be verified, and if they explicitly gave credit to all prior work they built upon.

Moreover, wealthy patrons and governments increasingly funded science as a profession, and required that findings be published in journals, thus maximally benefiting the public, as well as the public image of the patrons. This effectively created an economy based on \textit{reputation} \cite{kealey2010sex,nielsen2008future}. By publishing their findings, scientists were seen as trustworthy by their peers and patrons, which in turn led to better collaboration, research funding, and scientific jobs. This new culture continues to this day and has created a body of shared knowledge that is the basis for much of human progress.

Today, however, the ubiquity of the internet is allowing new, more scalable forms of scientific collaboration. We can now share detailed observations and methods (data and code) far beyond what can be printed in journals, and interact in real time with many people at once, all over the world. 

As a result, many sciences have turned to online tools to share, structure and analyse scientific data on a global scale. Such \textit{networked science} is dramatically speeding up discovery because scientists are now capable to build directly on each other's observations and techniques, reuse them in unforeseen ways, mine all collected data for patterns, and scale up collaborations to tackle much harder problems. Whereas the journal system still serves as our collective long-term memory, the internet increasingly serves as our collective \textit{short-term working memory} \cite{nielsen2012reinventing}, collecting data and code far too extensive and detailed to be comprehended by a single person, but instead (re)used by many to drive much of modern science.

Many challenges remain, however. In the spirit of the journal system, these online tools must also ensure that shared data is trustworthy so that others can build on it, and that it is in individual scientists' best interest to share their data and ideas.

In this paper, we discuss how other sciences have succeeded in building successful networked science tools that led to important discoveries, and build on these examples to introduce OpenML, a collaboration platform through which scientists can automatically share, organize and discuss machine learning experiments, data, and algorithms.

First, we explore how to design networked science tools in Section \ref{networkedscience}. Next, we discuss why networked science would be particularly useful in machine learning in Section \ref{related}, and describe OpenML in Section \ref{openml}. In Section \ref{benefits}, we describe how OpenML benefits individual scientists, students, and machine learning research as a whole, before discussing future work in Section \ref{future}. Section \ref{conclusions} concludes.

\section{Networked science}
\label{networkedscience}
Networked science tools are changing the way we make discoveries in several ways. They allow hundreds of scientists to discuss complex ideas online, they structure information from many scientists into a coherent whole, and allow anyone to reuse all collected data in new and unexpected ways. In this section we discuss how to design such online tools, and how several sciences have used them to make important breakthroughs.

\subsection{Designing networked science}
\label{design}
Nielsen \cite{nielsen2012reinventing} reviews many examples of networked science, and explains their successes by the fact that, 
through the interaction of many minds, there is a good chance that someone has just the right expertise to contribute at just the right time:
\begin{description} 
\item[Designed serendipity] Because many scientists have complementary expertise, any shared idea, question, observation, or tool may be noticed by someone who has just the right (micro)expertise to spark new ideas, answer questions, reinterpret observations, or reuse data and tools in unexpected new ways. By scaling up collaborations, such `happy accidents' become ever more likely and frequent.
\item[Dynamic division of labor] Because each scientist is especially adept at certain research tasks, such as generating ideas, collecting data, mining data, or interpreting results, any seemingly hard task may be routine for someone with just the right skills, or the necessary time or resources to do so. This dramatically speeds up progress.
\end{description}


Designed serendipity and a dynamic division of labor occur naturally when ideas, questions, data, or tools are broadcast to a large group of people in a way that allows everyone in the collaboration to discover what interests them, and react to it easily and creatively. As such, for online collaborations to scale, online tools must make it practical for anybody to join and contribute any amount at any time. This can be expressed in the following `design patterns' \cite{nielsen2012reinventing}:

\begin{itemize}[noitemsep]
\item Encourage small contributions, allowing scientists to contribute in (quasi) real time. This allows many scientists to contribute, increasing the cognitive diversity and range of available expertise.
\item Split up complex tasks into many small subtasks that can be attacked (nearly) independently. This allows many scientists to contribute individually and according to their expertise.
\item Construct a rich and structured information commons, so that people can efficiently build on prior knowledge. It should be easy to find previous work, and easy to contribute new work to the existing body of knowledge.
\item Human attention doesn't scale infinitely. Scientists only have a limited amount of attention to devote to the collaboration, and should thus be able to focus on their interests and filter out irrelevant contributions.
\item Establish accepted methods for participants to interact and resolve disputes. This can be an `honor code' that encourages respectable and respectful behavior, deters academic dishonesty, and protects the contributions of individual scientists.
\end{itemize}

Still, even if scientists have the right expertise or skill to contribute at the right time, they typically also need the right incentive to do so. 


As discussed, scientists actually solved this problem centuries ago by establishing a reputation system implemented using the best medium for sharing information of the day, the journal. Today, the internet and networked science tools provide a much more powerful medium, but they also need to make sure that sharing data, code and ideas online is in scientists' best interest.

The key to do this seems to lie in extending the reputation system \cite{nielsen2012reinventing}. Online tools should allow everyone to see exactly who contributed what, and link valuable contributions to increased esteem amongst the users of the tools and the scientific community at large. The traditional approach to do this is to link useful online contributions to authorship in ensuing papers, or to link the reuse of shared data to citation of associated papers or DOI's.\footnote{Digital Object Identifiers can be cited in papers. See, for instance, DataCite (http://www.datacite.org).} 

Moreover, beyond bibliographic measures, online tools can define new measures to demonstrate the scientific (and societal) impact of contributions. These are sometimes called \textit{altmetrics} \cite{altmetrics} or article-level metrics\footnote{http://article-level-metrics.plos.org/alm-info/}. An interesting example is ArXiv\footnote{http://arxiv.org}, an online archive of preprints (unpublished manuscripts) with its own reference tracking system (SPIRES). In physics, preprints that are referenced many times have a high status among physicists. They are added to resumes and used to evaluate candidates for scientific jobs. This illustrates that what gets measured, gets rewarded, and what gets rewarded, gets done \cite{nielsen2012reinventing,ostrom2000collective}. If scholarly tools define useful new measures and track them accurately, scientists will use them to assess their peers.

\subsection{Massively collaborative science}
Online tools can scale up scientific collaborations to any number of participants. In mathematics, Fields medalist Tim Gowers proposed\footnote{http://gowers.wordpress.com/2009/01/27/is-massively-collaborative-mathematics-possible} to solve 
several problems that have eluded mathematicians for decades by uniting many minds in an online discussion. Each of these \textit{Polymath projects} state a specific, unsolved math problem, is hosted on a blog\footnote{http://polymathprojects.org} or wiki\footnote{http://michaelnielsen.org/polymath1}, and invites anybody who has anything to say about the problem to chip in by posting new ideas and partial progress.

Designed serendipity plays an important role here. Each idea, even if just a hunch, may spark daughter ideas with those who happen to have just the right background. Indeed, several polymaths ``found themselves having thoughts they would not have had without some chance remark of another contributor''.\footnote{http://gowers.wordpress.com/2009/03/10/polymath1-and-open-collaborative-mathematics/} There is also a clear dynamic division of labor, with many mathematicians throwing out ideas, criticizing them, synthesizing, coordinating, and reformulating the problem to different subfields of mathematics. 

Blogs and wikis are ideally suited as tools, because they are designed to scale up conversations. They ensure that each contribution is clearly visible, stored and indexed, so that anybody can always see exactly what and how much you contributed.\footnote{Similarly, open source software development tools also allow anyone to see who contributed what to a project.} Moreover, everyone can make quick, small contributions by posting comments, all ideas are organized into threads or pages, and new threads or pages can be opened to focus on subproblems. In addition, anybody can quickly scan or search the whole discussion for topics of interest. 

Protected by a set of ground rules, individual scientists also receive rewards for sharing their ideas:

\begin{description}
\item[Authorship] Each successful polymath project resulted in several papers, linked to the original discussion. Via a self-reporting process, participants put forward their own names for authorship if they made important scientific contributions, or to be mentioned in the acknowledgements for smaller contributions.
\item[Visibility] Making many useful contributions may earn you the respect of notable peers, which is valuable in future collaborations or grant and job applications.
\item[Productivity] A scientist's time and attention is limited. It is profoundly enjoyable to contribute to many projects where you have a special insight or advantage, while the collaboration dynamically picks up other tasks.
\item[Learning] Online discussions are very engaging. Nascent ideas are quickly developed, or discarded, often leading to new knowledge or insight into the thought patterns of others.You are also encouraged to share an idea before someone else gets the same idea.
\end{description}

\subsection{Open data}
Online tools also collect and organize massive amounts of scientific data which can be mined for interesting patterns. For instance, the Sloan Digital Sky Survey (SDSS) is a collaboration of astronomers operating a telescope that systematically maps the sky, producing a stream of photographs and spectra that currently covers more than a quarter of the sky and more than 930,000 galaxies.\footnote{http://skyserver.sdss3.org} Although for a limited time, the data is only available to members of the collaboration, the SDSS decided to share it afterwards with the entire worldwide community of astronomers through an online interface \cite{szalay2002sdss}.\footnote{The data is also used in Microsoft's WorldWide Telescope (http://www.worldwidetelescope.org) and Google Sky (http://www.google.com/sky).} Since then, thousands of new and important discoveries have been made by analysing the data in many new ways \cite{boroson2009candidate}.

These discoveries are again driven by designed serendipity. Whereas the Polymath projects broadcast a question hoping that many minds may find a solution, the SDSS broadcasts data in the belief that many minds will ask unanticipated questions that lead to new discoveries. Indeed, because the telescope collects more data than a single person can comprehend, it becomes more of a question of asking the right questions than making a single `correct' interpretation of the data. Moreover, there is also a clear dynamic division of labor: the astronomers who ask interesting questions, the SDSS scientists who collect high quality observations, and the astroinformaticians who mine the data all work together doing what they know best.

Moreover, making the data publicly available is rewarding for the SDSS scientists are well.
\begin{description}
\item[Citation] Publishing data openly leads to more citation because other scientists can more easily build on them. In this case, other astronomers will use it to answer new questions, and credit the SDSS scientists. In fact, each data release easily collects thousands of citations. 
\item[Funding] Sharing the data increases the value of the program to the community as a whole, thus making it easier to secure continued funding. Indeed, if the data was not shared, reviewers may deem that the money is better spent elsewhere \cite{nielsen2012reinventing}. In fact, journals and grant agencies are increasingly expecting all data from publicly funded science to be published after publication.
\end{description}

The evolution towards more open data is not at all limited to astronomy.  We are `mapping and mining' just about every complex phenomenon in nature, including the brain \cite{lein2007genome,hawrylycz2012anatomically}, the ocean \cite{isern2006ocean}, gene sequences, genetic variants in humans \cite{frazer2007second}, and gene functions \cite{parkinson2007arrayexpress}. In many of these projects, the data is produced piece by piece by many different scientists, and gathered in a central database which they all can access.

\subsection{Citizen science}
Online tools are also enhancing the relationship between science and society. In \textit{citizen science} \cite{raddick2010galaxy}, the public is actively involved in scientific research. One example is Galaxy Zoo \cite{lintott2008galaxy}, where citizen scientists are asked to classify the galaxies from the SDSS and other sources such as the Hubble Space Telescope. Within a year, Galaxy Zoo received over 50 million classifications contributed by more than 150,000 people. These classifications led to many new discoveries, and the public data releases are cited hundreds of times.

Once again, designed serendipity occurs naturally. Unexpected observations are reported and discussed on an online forum, and have already resulted in the serendipitous discovery of the previously unknown `green pea' galaxies \cite{cardamone2009galaxy}, `passive red spirals' \cite{masters2010galaxy}, and other objects such as `Hanny's Object' \cite{lintott2009galaxy}, named after the volunteer who discovered it. Moreover, in a dynamic division of labor, citizen scientists take over tasks that are too time-consuming for professional astronomers. In fact, the overall classifications proved more accurate than the classifications made by a single astronomer, and obtained much faster. More engaged volunteers also participate in online discussions, or hunt for specific kinds of objects.

Galaxy Zoo and similar tools are also designed for scalability. The overall task is split up in many small, easy to learn tasks, each volunteer classifies as many galaxies as she wants, and classifications from different users are combined and organized into a coherent whole.

Finally, there are many different reasons for citizen scientists to dedicate their free time \cite{raddick2010galaxy}. Many are excited to contribute to scientific research. This can be out of a sense of discovery, e.g., being the first to see a particular galaxy, or because they believe in the goal of the project, such as fighting cancer. Many others view it as a game, and find it fun to classify many images. Some citizen science projects explicitly include a gamification component \cite{cooper2010predicting}, providing leaderboards and immediate feedback to volunteers. Finally, many volunteers simply enjoy learning more about a specific subject, as well as meeting new people with similar interests.

Citizen science is being employed in many more scientific endeavors\footnote{https://www.zooniverse.org/}, including 
protein folding \cite{cooper2010predicting}, planet hunting \cite{schwamb2012planet}, classifying plankton\footnote{http://www.planktonportal.org/}, and fighting cancer\footnote{http://www.cellslider.net}. Many of them are collecting large amounts of valid scientific data, and have yielded important discoveries.
 
\section{Machine learning}
\label{related}
Machine learning is a field where a more networked approach would be particularly valuable. Machine learning studies typically involve large data sets, complex code, large-scale evaluations and complex models, none of which can be adequately represented in papers. Still, most work is only published in papers, in highly summarized forms such as tables, graphs and pseudo-code. Oddly enough, while machine learning has proven so crucial in analysing large amounts of observations collected by other scientists, such as the SDSS data discussed above, the outputs of machine learning research are typically not collected and organized in any way that allows others to reuse, reinterpret, or mine these results to learn new things, e.g., which techniques are most useful in a given application.

\subsection{Reusability and reproducibility}
This makes us duplicate a lot of effort, and ultimately slows down the whole field of machine learning \cite{Vanschoren12,Hand:2006p14305}. Indeed, without prior experiments to build on, each study has to start from scratch and has to rerun many experiments. This limits the depth of studies and the interpretability and generalizability of their results \cite{Aha:1992p455,Hand:2006p14305}. It has been shown that studies regularly contradict each other because they are biased toward different datasets \cite{Keogh:2003p4930}, or because they don't take into account the effects of dataset size, parameter optimization and feature selection \cite{Perlich:2003p12674,Hoste:2005p12719}. This makes it very hard, especially for other researchers, to correctly interpret the results. Moreover, it is often not even possible to rerun experiments because code and data are missing, or because space restrictions imposed on publications make it practically infeasible to publish many details of the experiment setup. This lack of reproducibility has been warned against repeatedly \cite{Keogh:2003p4930,Sonnenburg:2007p12564,Pedersen:2008p12980}, and has been highlighted as one of the most important challenges in data mining research \cite{Hirsh:2008p14360}.


\subsection{Prior work}
Many machine learning researchers are well aware of these issues, and have worked to alleviate them. To improve reproducibility, there exist repositories to publicly share benchmarking datasets, such as UCI \cite{Asuncion:2007p519}, LDC\footnote{http://www.ldc.upenn.edu} and mldata\footnote{http://mldata.org}. Moreover, software can be shared on the MLOSS website\footnote{http://mloss.org}. There also exists an open source software track in the Journal for Machine Learning Research (JMLR) where short descriptions of useful machine learning software can be submitted. Also, some major conferences have started checking submissions for reproducibility \cite{Manolescu:2008p18044}, or issue open science awards for submissions that are reproducible.\footnote{http://www.ecmlpkdd2013.org/open-science-award/}

Moreover, there also exist experiment repositories. First, meta-learning projects such as StatLog \cite{Michie:1994p1780} and MetaL \cite{Brazdil:2009p721}, and benchmarking services such as MLcomp\footnote{http://www.mlcomp.org} run many algorithms on many datasets on their servers. This makes benchmarks comparable, and even allows the building of meta-models, but it does require that code be rewritten to run on their servers. Moreover, the results are not organized to be easily queried and reused.

Second, data mining challenge platforms such as Kaggle \cite{Carpenter:2011p34283} and TunedIT \cite{Wojnarski:2010p18714} collect results obtained by different competitors. While they do scale and offer monetary incentives, they are adversarial rather than collaborative. For instance, code is typically not shared during a competition. 

Finally, we previously introduced the \textit{experiment database for machine learning} \cite{Blockeel:2007p66,Vanschoren12}, which organizes results from different users and makes them queryable through an online interface. Unfortunately, it doesn't allow collaborations to scale easily. It requires researchers to transcribe their experiments into XML, and only covers classification experiments. 

While all these tools are very valuable in their own right, and we will build on them in this paper, they fail many of the requirements for scalable collaboration discussed above. It can be quite hard for scientists to contribute, there is often no online discussion, and they are heavily focused on benchmarking, not on sharing other results such as models. 


\section{OpenML}
\label{openml}
OpenML\footnote{OpenML is available on http://www.openml.org} is a place where machine learning researchers can automatically share data in fine detail and organize it to work more effectively and collaborate on a global scale.

It allows anyone to challenge the community with new data to analyze, and everyone able to mine that data to share their code and results (e.g., models, predictions, and evaluations).\footnote{In this sense, OpenML is similar to data mining challenge platforms, except that it allows users to work collaboratively, building on each other's work.} OpenML makes sure that each (sub)task is clearly defined, and that all shared results are stored and organized online for easy access, reuse and discussion.

Moreover, OpenML links to data available anywhere online, and is being integrated \cite{ecmldemo} in popular data mining platforms such as Weka \cite{Hall:2009p14495}, R \cite{mlr,Torgo:2010:DMR:1951702}, MOA \cite{Bifet:2010p28524}, RapidMiner \cite{RCOMM2013} and KNIME \cite{Berthold:2008p28224}. This means that anyone can easily import the data into these tools, pick any algorithm or workflow to run, and automatically share all obtained results. The OpenML website provides easy access to all collected data and code, compares all results obtained on the same data or algorithms, builds data visualizations, and supports online discussions.

Finally, it is an open source project, inviting scientists to extend it in ways most useful to them.

\subsection{How OpenML works}
OpenML offers various services to share and find data sets, to download or create scientific \textit{tasks}, to share and find implementations (called \textit{flows}), and to share and organize results. These services are available through the OpenML website, as well as through a REST API for integration with software tools.\footnote{In this paper, we only discuss the web interfaces. API details can be found on http://www.openml.org/api}

\subsubsection{Data sets}
\begin{figure}
\centering
\includegraphics[width=\columnwidth]{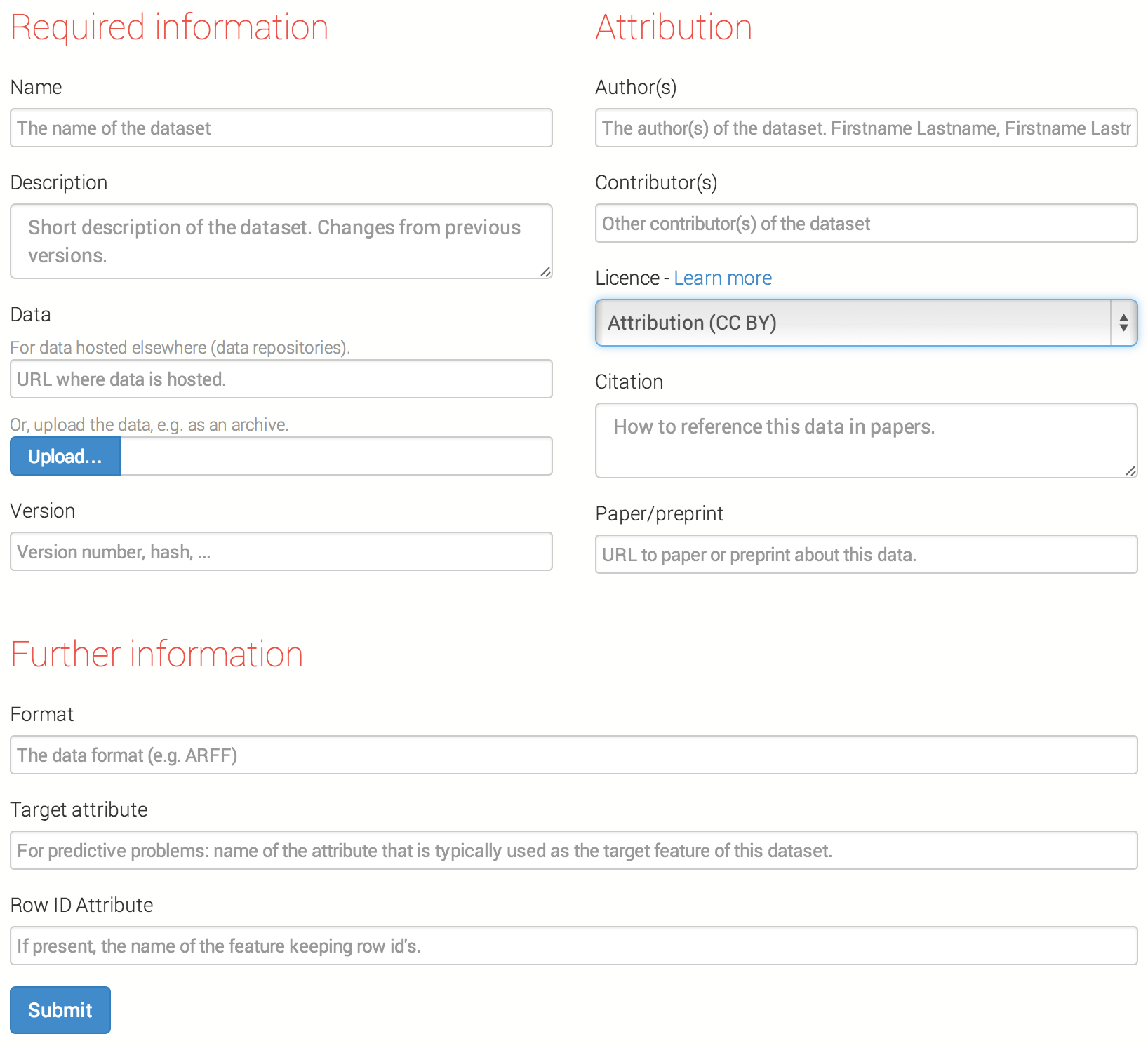}
\caption{\label{fig:data}Uploading data to OpenML.}
\end{figure}

Anyone can challenge the community with new data sets to analyze. Figure \ref{fig:data} shows how this is done through the website. To be able to analyse the data, OpenML accepts a limited number of formats. For instance, currently it requires the ARFF\footnote{http://www.cs.waikato.ac.nz/ml/weka/arff.html} format for tabular data, although more formats will be added over time.

The data can either be uploaded or referenced by a URL. This URL may be a landing page with further information or terms of use, or it may be an API call to large repositories of scientific data such as the SDSS.\footnote{In some cases, such as Twitter feeds, data may be dynamic, which means that results won't be repeatable. However, in such tasks, repeatability is not expected.} OpenML will automatically version each newly added data set. Optionally, a user-defined version name can be added for reference. Next, authors can state how the data should be attributed, and which (creative commons) licence they wish to attach to it. Authors can also add a reference for citation, and a link to a paper. Finally, extra information can be added, such as the (default) target attribute(s) in labeled data, or the row-id attribute for data where instances are named.

For known data formats, OpenML will then compute an array of  data characteristics. For tabular data, OpenML currently computes more than 70 characteristics\footnote{A full list can be found on http://www.openml.org/a}, including simple measures (e.g., the number of features, instances, classes, missing values), statistical and information-theoretic measures (e.g., skewness, mutual information) and landmarkers \cite{Pfahringer:2000p553}. Some characteristics are specific to subtypes of data, such as data streams. These characteristics are useful to link the performance of algorithms to data characteristics, or for meta-learning \cite{Vanschoren12} and algorithm selection \cite{LeiteBV12}.

OpenML indexes all data sets and allows them to be searched through a standard keyword search and search filters. Each data set has its own page with all known information.\footnote{See, for instance, http://www.openml.org/d/1} This includes the general description, attribution information, and data characteristics, but also statistics of the data distribution and, for each \emph{task} defined on this data (see below), all results obtained for that task. As will be discussed below, this allows you to quickly see which algorithms (and parameters) are best suited for analysing the data, and who achieved these results. It also includes a discussion section where the data set and results can be discussed.

\begin{figure}[t!]
\centering
\includegraphics[width=\columnwidth]{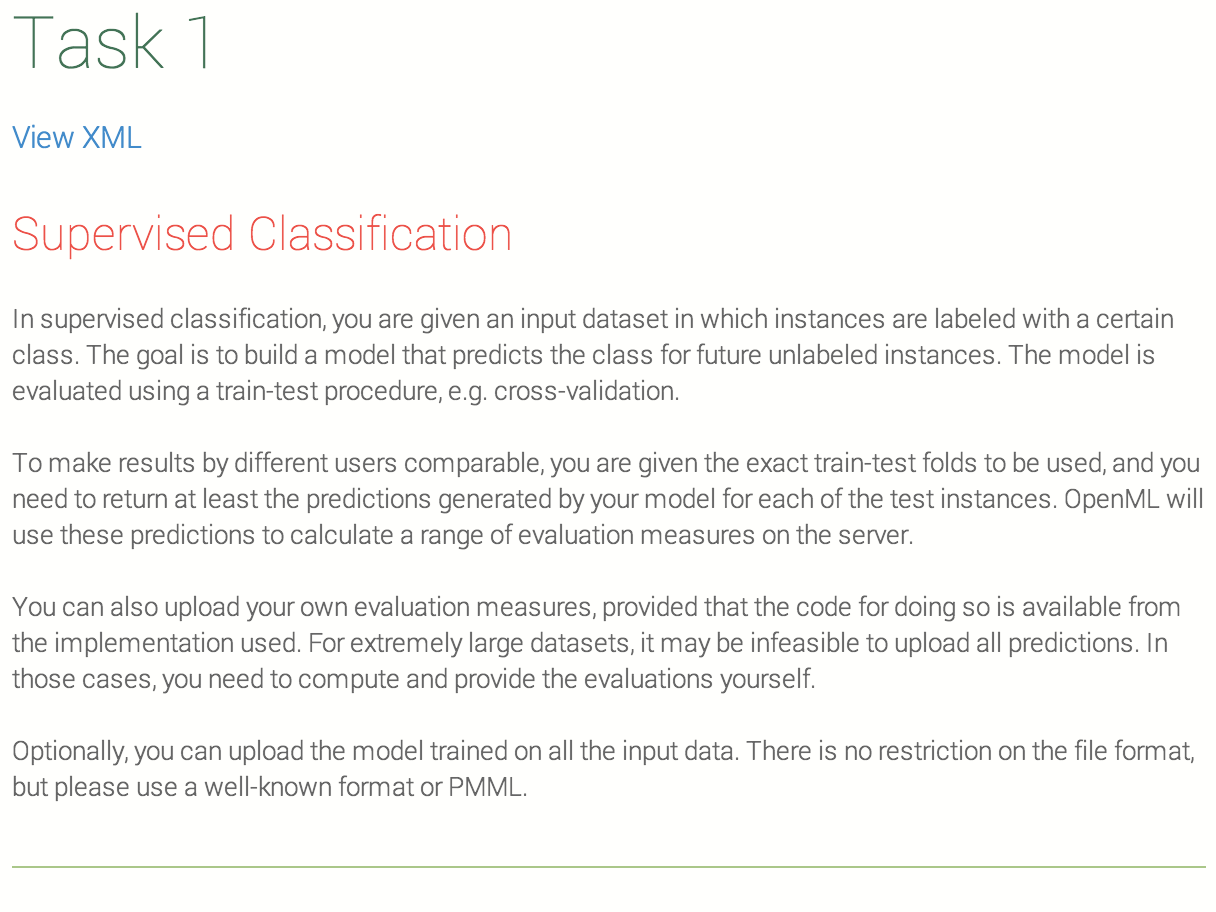}
\includegraphics[width=\columnwidth]{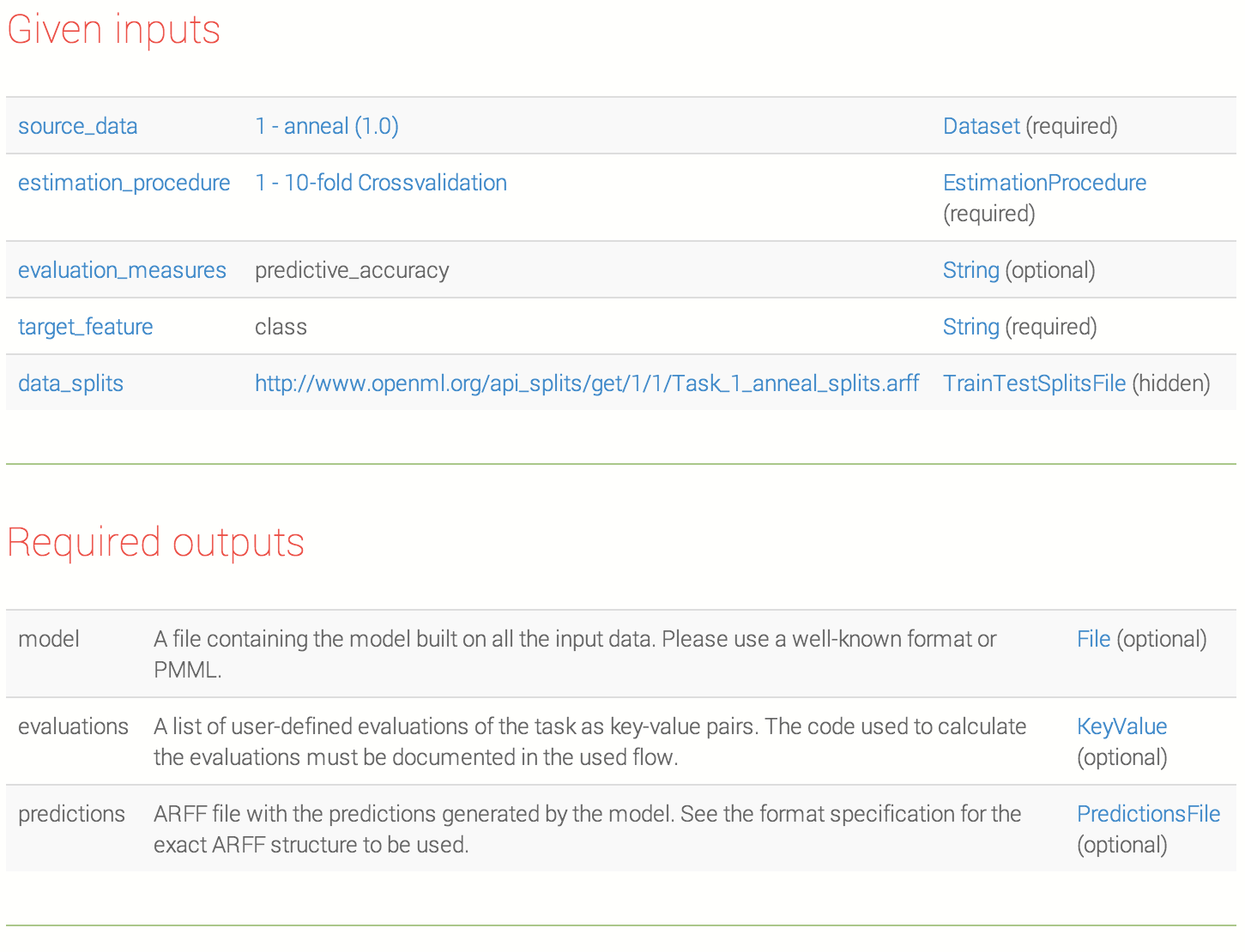}
\includegraphics[width=\columnwidth]{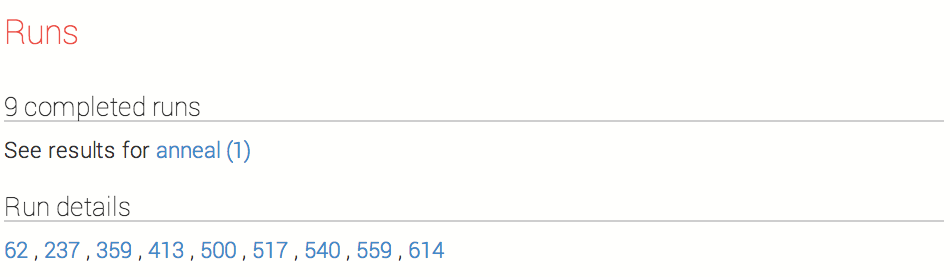}
\caption{\label{fig:task}An OpenML task (of task type  classification).}
\end{figure}

\subsubsection{Task types}
Obviously, a data set alone does not constitute a scientific challenge. We must first agree on what types of results are expected to be shared. This is expressed in \textit{task types}: they define what types of inputs are given, which types of output are expected to be returned, and what scientific protocols should be used. For instance, classification tasks should include well-defined cross-validation procedures, labeled input data, and require predictions as outputs.\footnote{Complete description: http://www.openml.org/t/type/1}

OpenML currently covers classification, regression, learning curve analysis and data stream classification. Task types are created by machine learning (sub)communities through the website, and express what they think should ideally be shared. In some cases, additional support may be required, such as running server-side evaluations. Such support will be provided upon request.
 
\subsubsection{Tasks}
If scientists want to perform, for instance, classification on a given data set, they can create a new machine learning \textit{task} online. Tasks are instantiations of task types with specific inputs (e.g., data sets). Tasks are created once, and then downloaded and solved by anyone. 

Such a task is shown in Figure \ref{fig:task}. In this case, it is a classification task defined on data set `anneal' version 1. Next to the data set, the task includes the target attribute, the evaluation procedure (here: 10-fold cross-validation) and a file with the data splits for cross-validation. The latter ensures that results from different researchers can be objectively compared. For researchers doing an (internal) hyperparameter optimization, it also states the evaluation measure to optimize for. The required outputs for this task are the predictions for all test instances, and optionally, the models built and evaluations calculated by the user. However, OpenML will also compute a large range of evaluation measures on the server to ensure objective comparison.\footnote{The evaluation measures and the exact code can be found on http://www.openml.org/a.}

Finally, each task has its own numeric id, a machine-readable XML description, as well as its own web page including all runs uploaded for that task, see Figure \ref{fig:task}.

\subsubsection{Flows}
\begin{figure}
\centering
\includegraphics[width=\columnwidth]{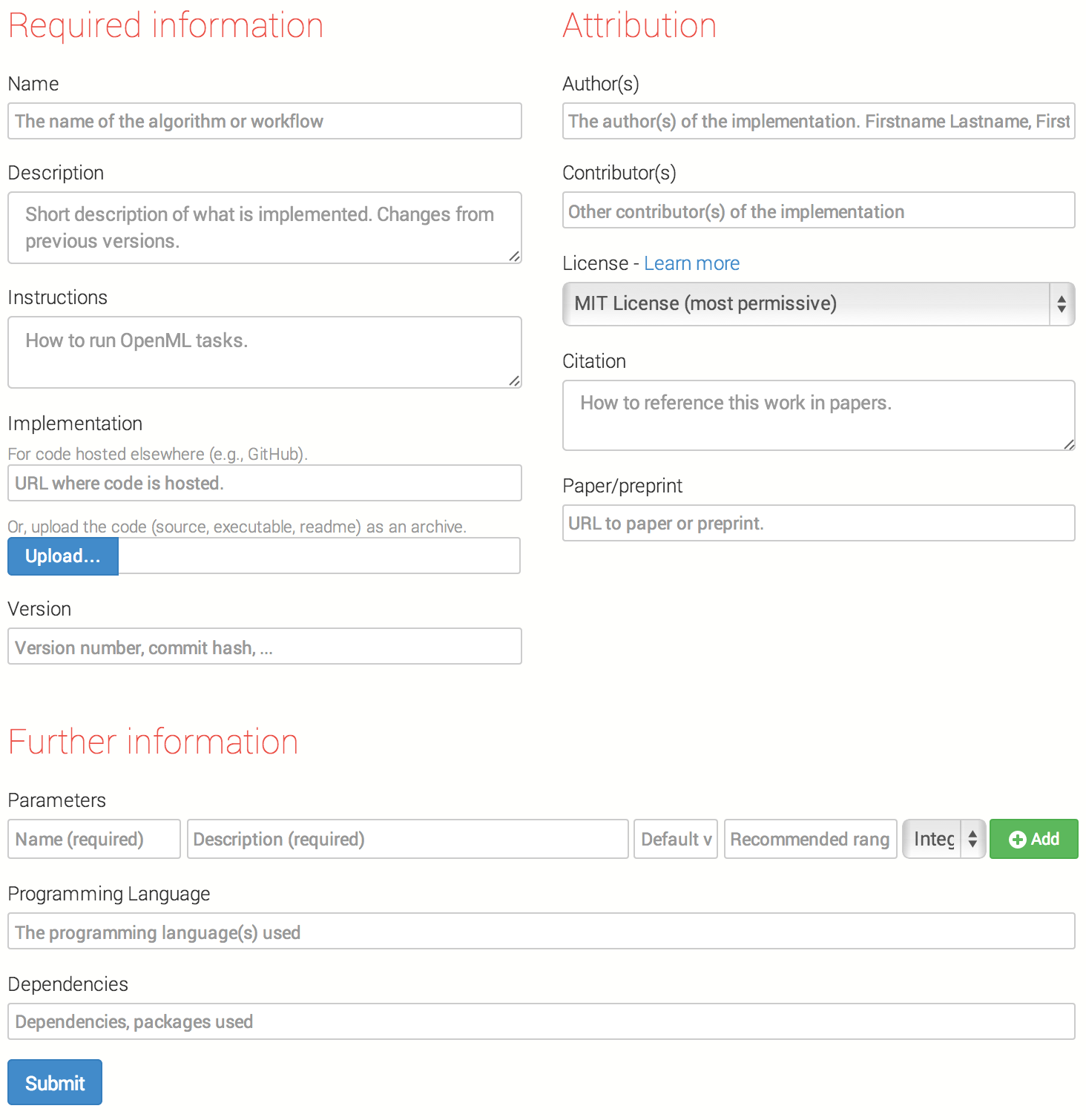}
\caption{\label{fig:flows}Uploading flows to OpenML.}
\end{figure}

\textit{Flows} are implementations of single algorithms, workflows, or scripts designed to solve a given task. They are uploaded to OpenML as shown in Figure \ref{fig:flows}. Again, one can upload the actual code, or reference it by URL. The latter is especially useful if the code is hosted on an open source platform such as GitHub or CRAN. Flows can be updated as often as needed. OpenML will again version each uploaded flow, while users can provide their own version name for reference. Ideally, what is uploaded is software that takes a task id as input and then produces the required outputs. This can be a wrapper around a more general implementation. If not, the description should include instructions detailing how users can run an OpenML task (e.g., to verify submitted results).

Attribution information is similar to that provided for data sets, although with a different set of licences. Finally, it is encouraged to add descriptions for the (hyper)parameters of the flow, and a range of recommended values.

\begin{figure}[!t]
\centering
\includegraphics[width=\columnwidth]{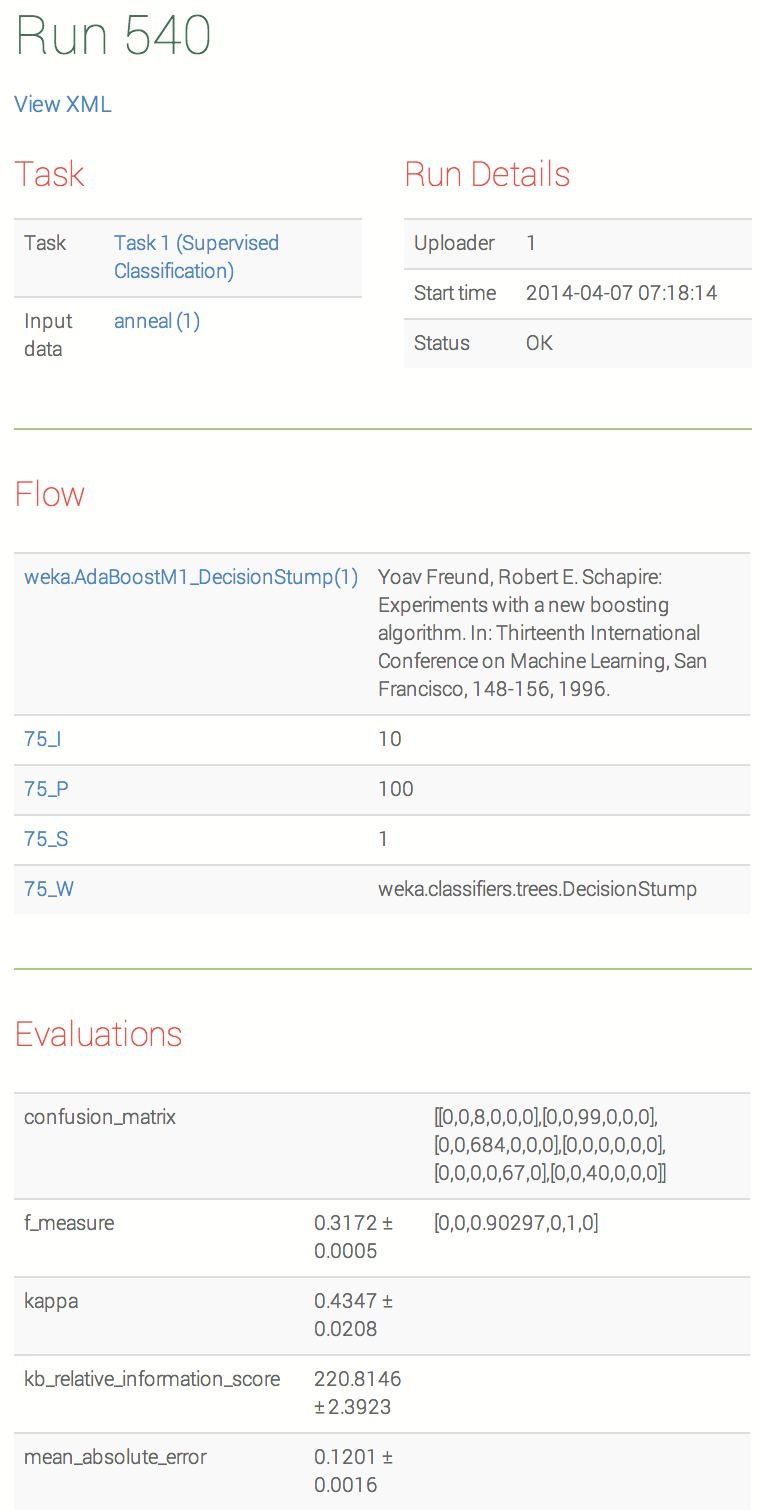}
\caption{\label{fig:run}An OpenML run.}
\end{figure}

It is also possible to annotate flows with characteristics, such as whether it can handle missing attributes, (non)numeric features and (non)numeric targets. As with data sets, each flow has its own page which combines all known information and all results obtained by running the flow on OpenML tasks, as well as a discussion section.

\subsubsection{Runs}
\textit{Runs} are applications of flows on a specific task. They are submitted by uploading the required outputs (e.g. predictions) together with the task id, the flow id, and any parameter settings. There is also a flag that indicates whether these parameter settings are default values, part of a parameter sweep, or optimized internally. Each run also has its own page with all details and results, shown partially in Figure \ref{fig:run}. In this case, it is a classification run. Note that OpenML stores the distribution of evaluations per fold (shown here as standard deviations), and details such as the complete confusion matrix and per-class results. We plan to soon add graphical measures such as ROC curves. Runtimes and details on hardware are provided by the user. 

\begin{figure}
\centering
\includegraphics[width=\columnwidth]{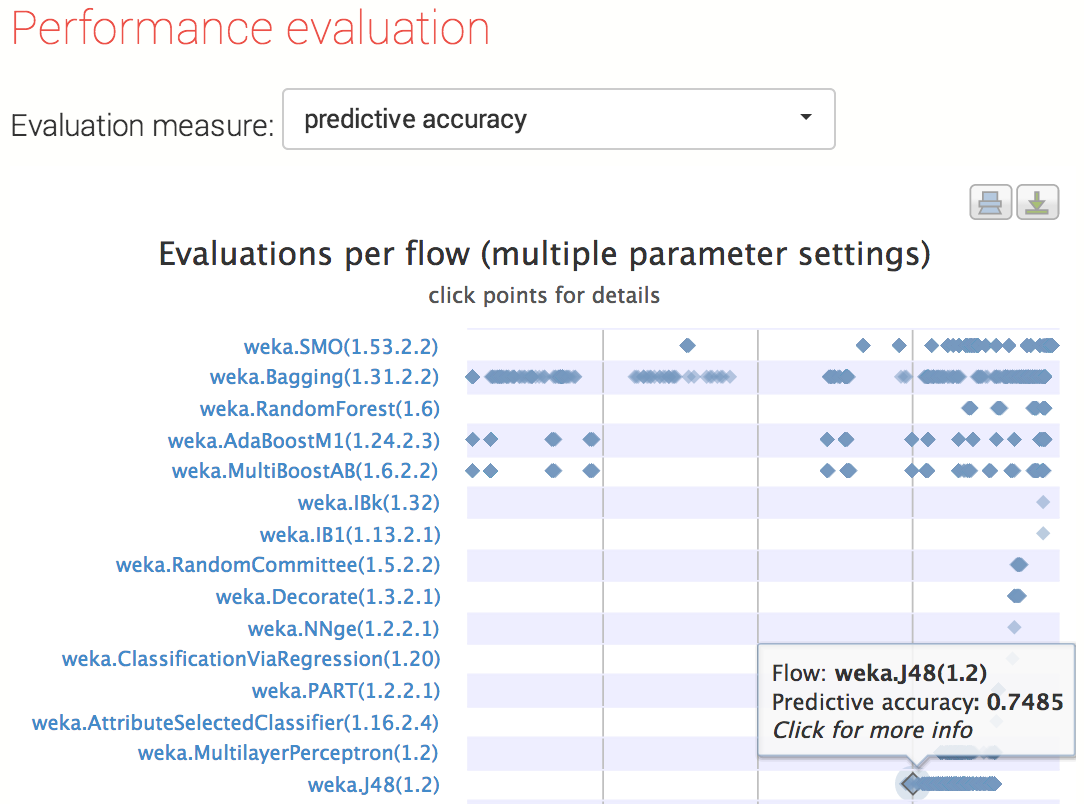}
\caption{\label{fig:datacompare} Portion of the page for data set `anneal'. It compares, for a classification task, the results obtained by different flows, for multiple parameter settings.}
\end{figure}

Moreover, because each run is linked to a specific task, flow, and author, OpenML will aggregate and visualize results accordingly. 

For instance, Figure \ref{fig:datacompare} shows a comparison of results obtained on a specific classification task. Each row represents a flow and each dot represents the performance obtained in a specific run (for different parameter settings). Hovering over a dot reveals more information, while clicking on it will pull up all information about the run. Users can also switch between different performance metrics.

\begin{figure}
\centering
\includegraphics[width=\columnwidth]{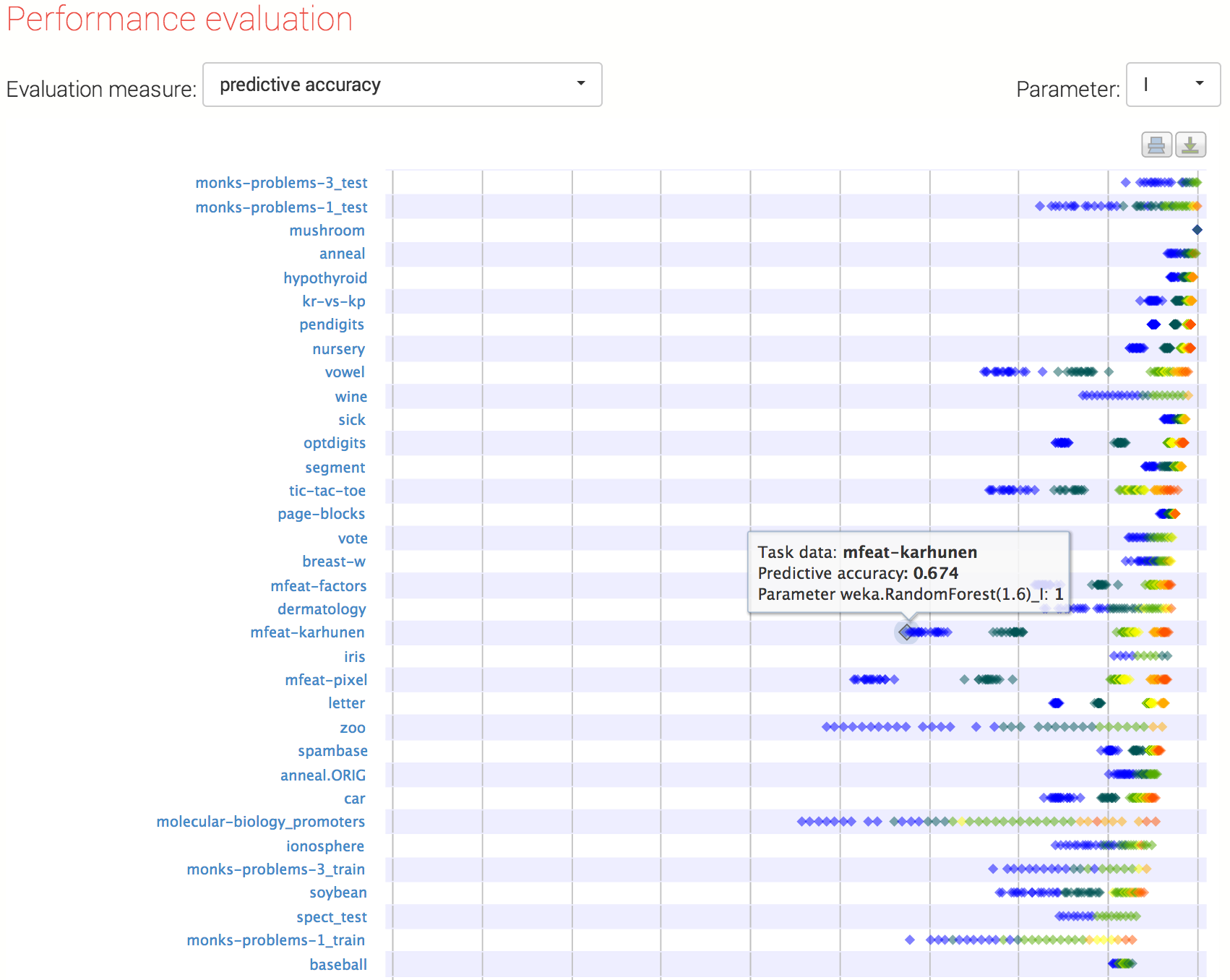}
\caption{\label{fig:compare} Portion of the page for flow `weka.RandomForest'. It compares, for classification tasks on different data sets, the results obtained by this flow for different parameter settings. Here, colored by the number of trees in the forest.}
\end{figure}

Conversely, Figure \ref{fig:compare} shows a comparison of results obtained by a specific flow on all tasks it has run on. Each row represents a task (and data set), and each dot the obtained performance. Additionally, it is possible to color-code the results with parameter values. Here, it shows the number of trees used in a random forest classifier from small (blue, left) to large (red, right). Again, clicking each dot brings up all run details. As such, it is easy to find out when, how, and by whom a certain result was obtained.

OpenML also provides several other task-specific visualizations such as learning curves. Moreover, it provides an SQL endpoint so that users can (re)organize results as they wish by writing their own queries. All results can be downloaded from the website for further study, and all visualizations can be exported. The database can also be queried programmatically through the API.

\subsection{OpenML plugins}
\label{plugins}
\begin{figure}
\centering
\includegraphics[width=\columnwidth]{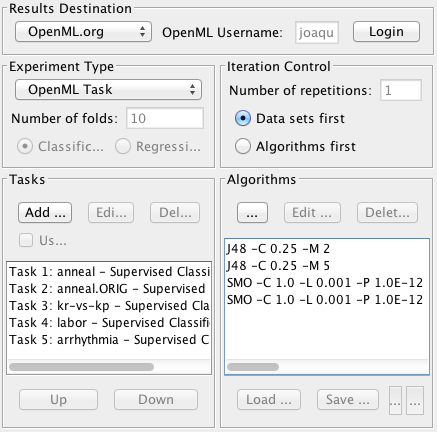}
\caption{\label{fig:weka}WEKA integration of OpenML.}
\end{figure}

\begin{figure}
\centering
\includegraphics[width=\columnwidth]{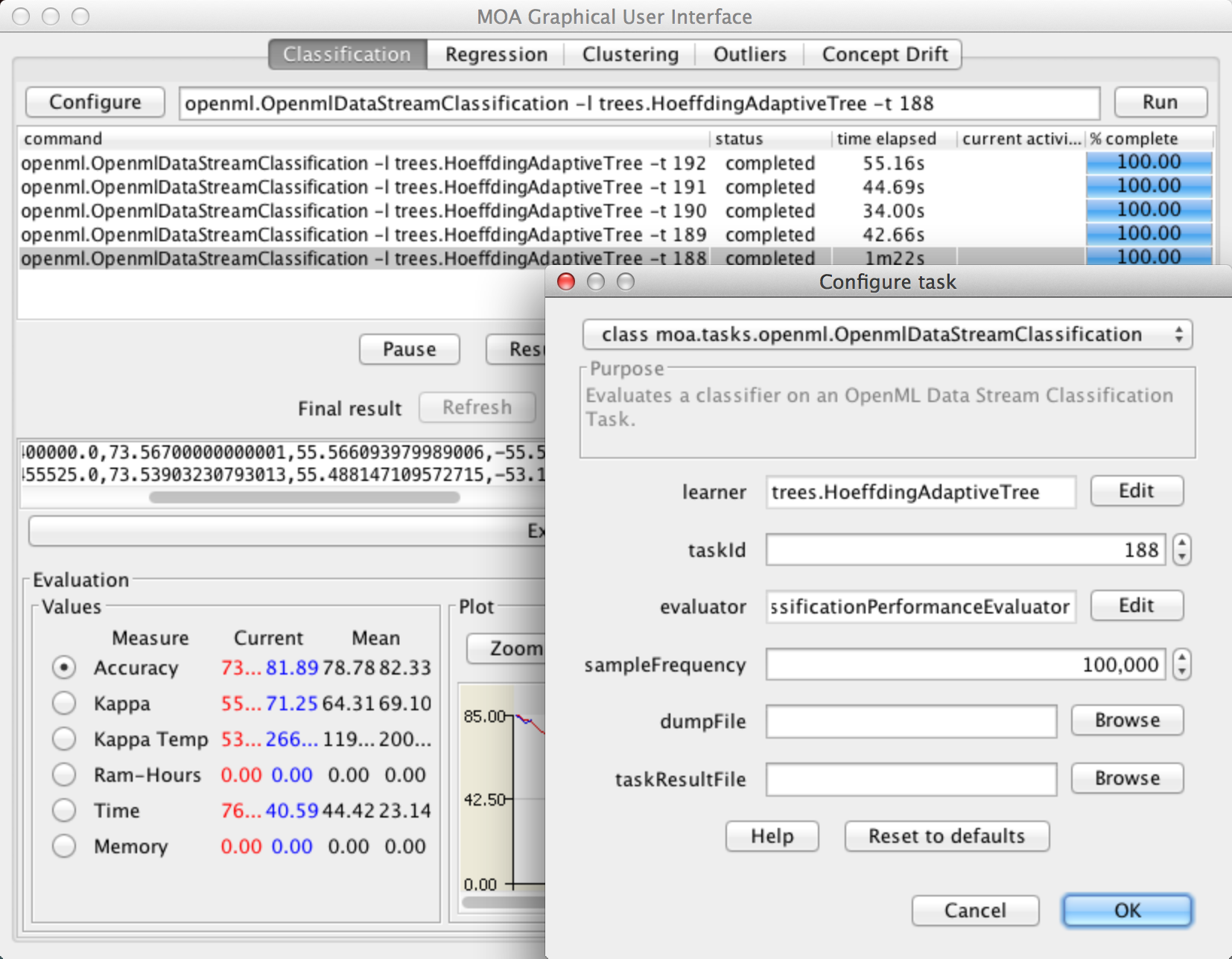}
\caption{\label{fig:moa}MOA integration of OpenML.}
\end{figure}

\begin{figure}
\centering
\includegraphics[width=\columnwidth]{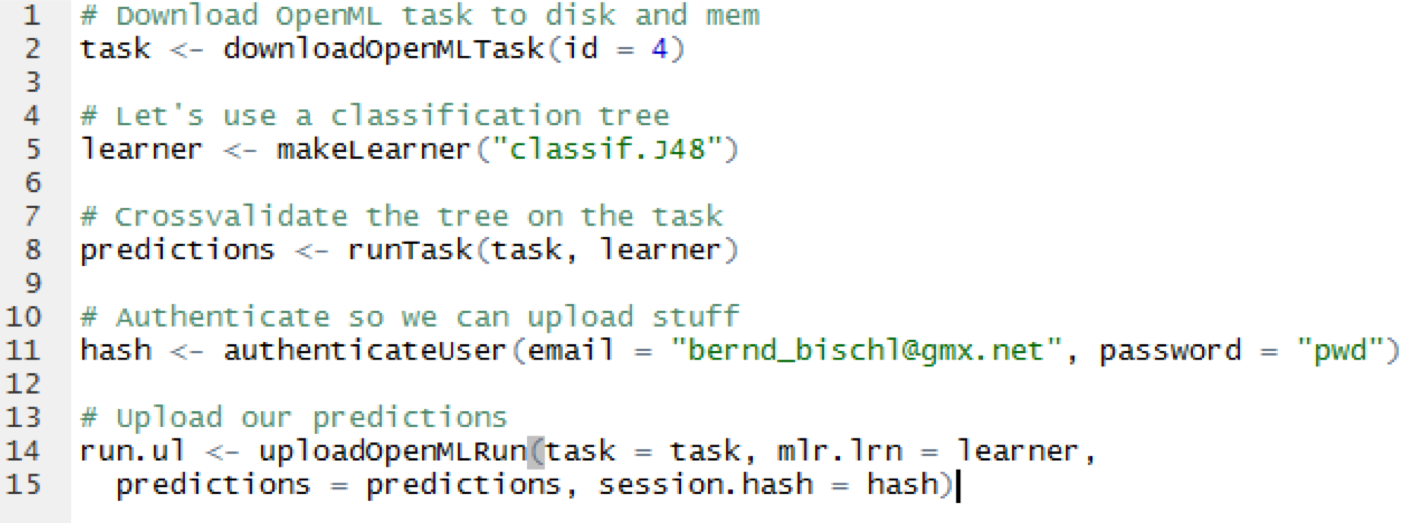}
\caption{\label{fig:r}R integration of OpenML.}
\end{figure}

As stated above, OpenML features an API so that scientific software tools can connect to it to download data or upload new results. However, existing tools can also connect to OpenML by simply adding a few lines of code, and without any knowledge of the API. Indeed, OpenML  provides language-specific libraries that take care of all server communication. For instance, software written in Java would use the OpenML Java interface to download tasks and upload results. More precisely, a method \textit{getTask(id)} will return a Java object containing all data to run the task, and a method \textit{submitRun(outputs)} will take the obtained results and submit them to OpenML. We also aim to provide command-line tools for connecting to OpenML.

On top of that, OpenML is being integrated in several popular machine learning environments, so that it can be used out of the box. These \textit{plugins} can be downloaded from the website. Figure \ref{fig:weka} shows how OpenML is integrated in WEKA's Experimenter \cite{Hall:2009p14495}. After selecting OpenML as the result destination and providing your login credentials, you can add a number of tasks through a dialogue (or simply provide a list of task id's), and add a number of WEKA algorithms to run. Behind the scenes, the plugin will download all data, run every algorithm on every task (if possible) and automatically upload the results to OpenML. The results will also be locally available in WEKA for further analysis.

For data stream mining, one can use the MOA \cite{Bifet:2010p28524} plugin as shown in Figure \ref{fig:moa}. Similarly to WEKA, users can select OpenML tasks, and then run any algorithm on them while uploading all runs to OpenML in the background.

Finally, researchers that use R can use the \textit{openml} package as shown in Figure \ref{fig:r}. One first downloads a task given a task id, then runs the task by providing a learner, and finally uploads the run by providing the task, learner, results and user authentication. While we do plan to integrate OpenML into other environments as well, at the time of writing these are still in development. 

\newpage
\section{Networked machine learning}
\label{benefits}
Through OpenML, we can initiate a fully networked approach to machine learning. In this section we compare OpenML to the networked science tools described before, and describe how it helps scientists make new discoveries, how it allows collaborations to scale, and how it benefits individual scientists, students and a more general audience.

\subsection{OpenML and networked science}
\label{promise}
By sharing and organizing machine learning data sets, code and experimental results at scale, we can stimulate designed serendipity and a dynamic division of labor.

\subsubsection{Designed serendipity}
Similar to the SDSS, by organizing and `broadcasting' all data, code and experiments, many minds may reuse them in novel, unforeseen ways. 

First, new discoveries could by made simply by \textit{querying} all combined experiments to answer interesting questions. These question may have been nearly impossible to answer before, but are easily answered if a lot of data is already available. In addition, with readily available data, it becomes a routine part of research to answer questions such as ``What is the effect of data set size on runtime?'' or ``How important is it to tune hyperparameter P?'' With OpenML, we can answer these questions in minutes, instead of having to spend days setting up and running new experiments \cite{Vanschoren12}. This means that more such questions will be asked, possibly leading to more discoveries.

Second, we can \textit{mine} all collected results and data characteristics for patterns in algorithm performance. Such \textit{meta-learning} studies could yield insight into which techniques are most suited for certain applications, or to better understand and improve machine learning techniques \cite{Vanschoren12}.

\begin{figure}[b]
\centering
\includegraphics[width=\columnwidth]{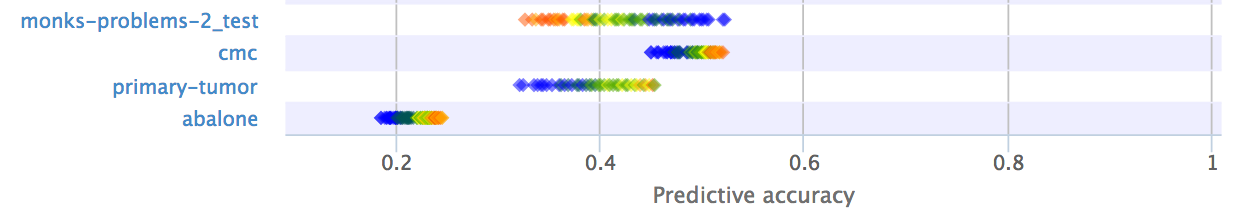}
\caption{\label{fig:unexpected} An unexpected result: the performance of a random forest decreases as more trees are added to the forest.}
\end{figure}

Third, anyone could run into unexpected results by browsing through all collected data. An example of this is shown in Figure \ref{fig:unexpected}, which is a continuation of the results shown in Figure \ref{fig:compare}: while the performance of a random forest classifier should increase (or stagnate) when more trees are added to the forest (red dots), it sometimes happens that it decreases. As in Galaxy Zoo, such serendipitous discoveries can be discussed online, combining many minds to explore several possible explanations. 

Finally, beyond experiments, data sets and flows can also be reused in novel ways. For instance, an existing technique may prove extremely useful for analysing a new data set, bringing about new applications.

\subsubsection{Dynamic division of labor}
\label{proofdynamic}
OpenML also enables a dynamic division of labor: large-scale studies could be undertaken as a team, or hard questions could be tackled collaboratively, with many scientists contributing according to their specific skills, time or resources. 

Scientists, possibly from other domains, can focus the attention of the community on an important problem. This can be done by adding new data sets (and tasks) to OpenML and collaborating with the machine learning community to analyse it. Some may suggest techniques that would otherwise not be considered, while others are especially skilled at designing custom-built workflows, running large-scale experiments, or improving code. Conversely, the scientists that contributed the data can provide direct feedback on the practical utility of suggested approaches, interpret the generated models, and otherwise guide the collaboration to the desired outcome. Such collaborations can scale to any number of scientists. OpenML also helps to coordinate the effort, e.g., by organizing all results per task (see Figure \ref{fig:compare}), so that everybody can track each other's progress, and discuss ideas and results online. 

Another case is that of benchmark studies. While it is important that a new algorithm be compared against the state of the art, it is often time-consuming to hunt down their implementations and to figure out how to run them. On OpenML, each researcher that invents a new algorithm can focus on experimenting with that algorithm alone, knowing best how to apply it on different tasks. Next, she can instantly reuse the results from all other shared algorithms, ran by their original authors. As such, a very complete overview of the state of the art emerges spontaneously.

Finally, students and citizen scientists can also contribute to research simply by using the OpenML plugins to experiment while they learn about machine learning techniques.

\subsection{Scaling up collaboration}
\label{tool}
As discussed in section \ref{design}, these benefits emerge faster if online collaborations are allowed to scale.

First of all, it is \textit{easy to make small contributions} to OpenML. When using any of the OpenML plugins, you can easily import a task, run any algorithm or workflow, and automatically export the results to OpenML. You can just perform a single run, a few, or thousands without much effort. Moreover, scientists who know of new interesting data sets or algorithms can easily add them through the website, and watch how others start experimenting with them. It is even easier to browse through the discussions running on OpenML, and leave a comment or suggestion. Alternatively, one can browse the results shared on the website, and draw attention to unexpected results that are worth investigating. 

Even contributing a single run, data set or comment can be valuable to the community, and may stimulate more work in that direction. More committed scientists can contribute in many other ways, such as creating new tasks and task types, adding new data characterizations or evaluation measures, and integrating OpenML in new tools and environments.

Moreover, OpenML tasks naturally \textit{split up complex studies} into tasks which can be run independently by many scientsists according to their skills, as discussed in Section \ref{promise}. Tasks also split the machine learning community into smaller subcommunities (e.g., clustering) which focus on a single task type, or subgroups focusing on a single task (e.g. galaxy clustering). Designed serendipity and dynamic division of labor also occur in small but active subcommunities. They are not held back if other communities are less active.

Next, OpenML constructs a \textit{rich and structured information commons}, building a database of all data sets, flows, tasks, runs, results, scientists, and discussions. OpenML also aggregates results in different ways, e.g., visualizing results per task and flow (see Figures \ref{fig:datacompare} and \ref{fig:compare}). Keyword searches and filters make it easy to find resources, and more complex questions can be answered through the SQL interface, or by downloading data and analysing it using other tools.

As a result, all information is also \textit{open but easily filtered}. The website organizes all results per task, data set and flow, so that researchers can focus on what interests them most, without being distracted by the activity of other scientists. In future work, we also aim to filter results by their authors.

Finally, OpenML establishes, and in some cases enforces, a scientific approach to sharing results. Indeed, OpenML tasks set a certain standard of scientific quality and trustworthiness by defining how experiments must be run and what must be reported. Because the code is shared when uploading runs, it is possible for others to verify results, and the server-side evaluations makes results objectively comparable. OpenML also makes clear who contributed what (and when), and how it is licenced. Every shared algorithm, flow, run or comment can be attributed to a specific person, and this information is always shown when someone views them online.

\subsection{Benefits for scientists}
\label{bargain}
How do you, as an individual scientist, benefit from sharing your experiments, data and code on OpenML?

\subsubsection{More time}
First, you gain more time. OpenML assists in most of the routine and tedious duties in running experiments: finding data sets, finding implementations, setting up experiments, and organizing all experiments for further analysis. Moreover, when running benchmark experiments on OpenML, you can directly compare them with the state of the art, reusing other, comparable results. In addition, you can answer routine research question in minutes by tapping into all shared data, instead of losing days setting up new experiments. Finally, having your experiments stored and organized online means they are available any place, any time, through any browser (including mobile devices), so you can access them when it is convenient.



\subsubsection{More knowledge}
Second, you gain more knowledge. Linking your results to everybody else's has a large potential for new discoveries. This was discussed in Section \ref{promise}: you can answer previously impossible questions, mine all combined data, and run into unexpected results. It also makes it easy to check whether certain observations in your data are echoed in the observations of others. Next, with OpenML you can interact with other minds on a global scale. Not only can you start discussions to answer your own questions, you can also help others, and in doing so, learn about other interesting studies, and forge new collaborations. Finally, by reusing prior results, you can launch larger, more generalizable studies that are practically impossible to run on your own.

\subsubsection{More reputation}
Third, OpenML helps you build reputation by making your work more visible to a wider group of people, by bringing you closer to new collaborators, and by making sure that others know how to credit you if they build on any of your work.
\begin{description}
\item[Citation] OpenML makes sure that all your contributions, every data set, flow and run, are clearly attributed to you. If others wish to reuse your results, OpenML will tell them how you wish to be credited (e.g., through citation). Moreover, OpenML makes your shared resources easy to find, thus making frequent citations more likely.
\item[Altmetrics] OpenML will also automatically track how often your data or code is reused in experiments (runs), and how often your experiments are reused in studies (see below). These are clear measures of the impact of your work.
\item[Productivity] OpenML allows you to contribute efficiently to many studies. This increases your scientific productivity, which translates to more publications.
\item[Visibility] You can increase your visibility by contributing to many studies, thus earning the respect of new peers. Also, if you design flows that outperform many others, OpenML will show these at the top of each data set page, as shown in Figure \ref{fig:datacompare}. You can also post links to your online results in blogs or tweets. 
\item[Funding] Open data sharing is increasingly becoming a requirement in grant proposals, and uploading your research to OpenML is a practical and convincing way to share your data with others.
\item[No publication bias] Most journals have a publication bias: even if the findings are valuable, it is hard to publish them if the outcome is not positive. Through OpenML, you can still share such results and receive credit for them.
\end{description}

\subsection{Benefits for students}
OpenML can also substantially help students in gaining a better understanding of machine learning. 
Browsing through organized results online is much more accessible than browsing through hundreds of papers. It provides a clear overview of the state of the art, interesting new techniques and open problems. As discussed before, students can contribute in small or big ways to ongoing research, and in doing so, learn more about how to become a machine learning researcher. Online discussions may point to new ideas or point out mistakes, so they can learn to do it better next time. In short, it gives students and young scientists a large playground to learn more quickly about machine learning and discover where they can make important contributions.

\section{Future work}
\label{future}
In this section, we briefly discuss some of the key suggestions that have been offered to improve OpenML, and we aim to implement these changes as soon as possible. In fact, as OpenML is an open source project, everyone is welcome to help extend it, or post new suggestions through the website.\\

\subsection{OpenML studies}
One typically runs experiments as part of a study, which ideally leads to a publication. Scientists should therefore be able to create \textit{online studies} on OpenML, that combine all relevant info on one page. Such studies reference all runs of interest, either generated for this study or imported from other OpenML studies, and all data sets and flows underlying these runs. Additionally, textual descriptions can be added to explain what the study is about, and any supplementary materials, such as figures, papers or additional data, can be uploaded and attached to it. 

If the study is published, a link to this online study can be added in the paper, so that people can find the original experiments, data sets and flows, and build on them. As such, it becomes the online counterpart of a published paper, and you could tell people to cite the published paper if they reuse any of the data. An additional benefit of online studies is that they can be extended after publication. 

Moreover, based on the underlying runs, OpenML can automatically generate a list of references (citations) for all underlying data sets, flows and other studies. This helps authors to properly credit data that they reused from other OpenML scientists. Similar to arXiv, OpenML can also automatically keep track of this so that authors can instantly view in which studies their contributions are being reused.

Finally, similar to the polymath projects, such studies could be massively collaborative studies, aimed at solving a hard problem, and driven by many people providing ideas, experiments, data or flows. As such, each study should have a discussion section where questions can be asked, suggestions can be made and progress can be discussed. Similar to the polymath studies, it also makes sense if studies link to other studies that tackle specific subproblems, and if the study was linked to a wiki page for collaborative writing.

To keep focus on the results of a given study, it can act as a filter, hiding all other results from the website so that only the contents of that study are visible.

\subsection{Visibility and social sharing}
There may be cases where you want all of OpenML's benefits, but do not want to make your data public before publication. On the other hand, you may want to share that data with trusted colleagues for collaboration or feedback, or allow friends to edit your studies, e.g., to add new runs to it. It therefore makes sense if, for a limited amount of time, studies, data sets or flows can be flagged as private or `friends only'. Still, scientists should agree that this is a temporary situation. When you wish to be attributed for your work, you must first make it publicly available. 

In addition, in highly experimental settings, most results may not be interesting as such. Scientists are always able to delete those results or mark them as deprecated.

\subsection{Collaborative leaderboards}
When many scientists work together to design a flow for a specific task, it is useful to show which flows are currently performing best, so that others can build on and improve those flows. However, it is also important to show which contributions had the biggest impact while such a flow was constructed collaboratively. In those cases, it is useful to implement a leaderboard that does not only show the current best solution, but instead credits the authors who contributed solutions that were in, say, the top 3 at any point in time. Alternatively, this can be visualized in a graph of performance versus time, so that it is clear who caused the bigger performance `jumps'. This is useful to later credit the people who made the most important contributions.



\subsection{Broader data support} 
To build a well-structured information space, OpenML needs to be able to correctly interpret the uploaded data. For instance, to calculate data characteristics and build train-test splits, more information is needed about the structure of data sets. Therefore, we have initially focused on ARFF data. However, many types of data, such as graphs, can not always be adequately expressed as ARFF, and we will add support for new data types according to researchers' needs. Moreover, for some types of tasks, additional types of results (run outputs) may need to be defined so that OpenML can properly interpret them. 

In the short term, we aim to add support for Graph Mining, Clustering, Recommender Systems, Survival Analysis, Multi-label Classification, Feature selection, Semi-Supervised Learning, and Text Mining. Moreover, we will extend the website to make it easy to propose and work collaboratively on support for new task types. 



\section{Conclusions}
\label{conclusions} 
In many sciences, networked science tools are allowing scientists to make discoveries much faster than was ever possible before. Hundreds of scientists are collaborating to tackle hard problems, individual scientists are building directly on the observations of all others, and students and citizen scientists are effectively contributing to real science.

To bring these same benefits to machine learning researchers, we introduce OpenML, an online service to share, organize and reuse data, code and experiments. Following best practices observed in other sciences, OpenML allows collaborations to scale effortlessly and rewards scientists for sharing their data more openly. 

We believe that this new, networked approach to machine learning will allow scientists to work more productively, make new discoveries faster, be more visible, forge many new collaborations, and start new types of studies that were practically impossible before. 

\section{Acknowledgements}
This work is supported by grant 600.065.120.12N150 from the Dutch Fund for Scientific Research (NWO), and by the IST Programme of the European Community, under the Harvest Programme of the PASCAL2 Network of Excellence, IST-2007-216886.

We also wish to thank Hendrik Blockeel, Simon Fisher and Michael Berthold for their advice and comments.

\bibliographystyle{abbrv}
\bibliography{sigproc}  
%
\end{document}